\documentclass[10pt,twocolumn,letterpaper]{article}

\usepackage{cvpr}              

\usepackage{graphicx}
\usepackage{amsmath}
\usepackage{amssymb}
\usepackage{booktabs}

\usepackage{amsfonts}
\usepackage{multirow}
\usepackage{mathrsfs}  
\usepackage{array}
\usepackage{tikz}
\usepackage{pgfplots}
\usepgfplotslibrary{groupplots}
\pgfplotsset{compat=1.17} 

\usepackage[acronym]{glossaries}
\newacronym{CNN}{CNN}{convolutional neural network}
\newacronym{FR}{FR}{face recognition}
\newacronym{GAN}{GAN}{generative adversarial network}
\newacronym{AAD}{AAD}{adaptive attentional denormalization}
\newacronym{AADRes}{AADRes}{\gls{AAD} residual block}
\newacronym{SPADE}{SPADE}{spatially adaptive normalization}
\newacronym{AdaIn}{AdaIn}{adaptive instance normalization}
\newacronym{LFW}{LFW}{labeled faces in the wild}

\DeclareMathOperator{\id}{id}
\newlength\figlen

\usepackage[pagebackref,breaklinks,colorlinks]{hyperref}

\usepackage[capitalize]{cleveref}
\crefname{section}{Sec.}{Secs.}
\Crefname{section}{Section}{Sections}
\Crefname{table}{Table}{Tables}
\crefname{table}{Tab.}{Tabs.}

\def\cvprPaperID{5} 
\def\confName{CVPR}
\def\confYear{2022}

\begin{document}

\title{Face Morphing: Fooling a Face Recognition System Is Simple!}

\author{
Stefan H\"ormann \quad Tianlin Kong \quad Torben Teepe \quad Fabian Herzog \quad Martin Knoche  \quad Gerhard Rigoll\\
Technical University of Munich\\
{\tt\small s.hoermann@tum.de}
}

\maketitle

\begin{abstract}
State-of-the-art \gls{FR} approaches have shown remarkable results in predicting whether two faces belong to the same identity, yielding accuracies between 92\% and 100\% depending on the difficulty of the protocol. However, the accuracy drops substantially when exposed to morphed faces, specifically generated to look similar to two identities. To generate morphed faces, we integrate a simple pretrained \gls{FR} model into a \gls{GAN} and modify several loss functions for face morphing. In contrast to previous works, our approach and analyses are not limited to pairs of frontal faces with the same ethnicity and gender. Our qualitative and quantitative results affirm that our approach achieves a seamless change between two faces even in unconstrained scenarios. Despite using features from a simpler \gls{FR} model for face morphing, we demonstrate that even recent \gls{FR} systems struggle to distinguish the morphed face from both identities obtaining an accuracy of only 55-70\%. Besides, we provide further insights into how knowing the \gls{FR} system makes it particularly vulnerable to face morphing attacks.
\end{abstract}
\glsresetall

\section{Introduction}

Drawing from the impressive results of \glspl{GAN}, face manipulation tasks have been investigated more frequently in the research community. Face manipulation is employed in multiple applications, \eg, face swapping \cite{korshunova2017fast, nirkin2019fsgan, li2019faceshifter}, face attribute manipulation \cite{he2019AttGAN}, face beautification \cite{Chen2019BeautyGlow, diamant2019BeholderGAN}, and (anti-)aging \cite{he2019AttGAN,Liu2021A3GAN, perov2021deepfacelab}. Face swapping targets substituting the identity of a face with the identity in a target image, maintaining background, head pose, and facial expression of the original image. Thus, identity features must be extracted and disentangled from the remaining information and introduced into the source image. 

\begin{figure}[t]
	\centering
	\includegraphics[width=\linewidth]{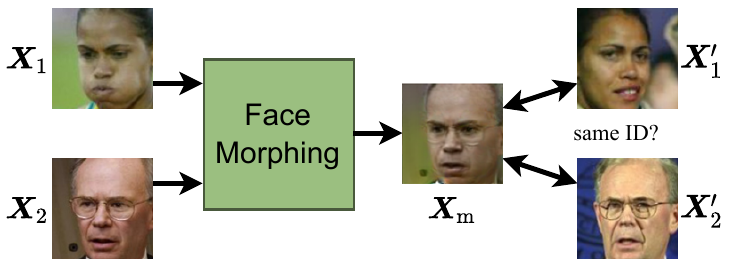}
	\caption{Example of a face morphing attack: The morphed face $\boldsymbol X_{\text{m}}$ is created given $\boldsymbol X_1$ and $\boldsymbol X_2$ from different identities. The objective is to determine whether an \gls{FR} system classifies $\boldsymbol X_{\text{m}}$ to have the same identities as $\boldsymbol X_1'$ and $\boldsymbol X_2'$ despite $\boldsymbol X_1'$ and $\boldsymbol X_2'$ being from different identities.}
	\label{fig:overview}
\end{figure}

In contrast to face swapping, face morphing aims to create a seamless transition between two faces, $\boldsymbol X_1$ and $\boldsymbol X_2$, which involves identity, attributes, head pose, and background. Hence, when considering the information from both faces equally, the morphed face $\boldsymbol X_{\text{m}}$ looks similar to $\boldsymbol X_1$ and $\boldsymbol X_2$, as depicted in \cref{fig:overview}. To ensure that security-sensitive applications such as automatic border control or access control are not exposed to morphed faces, the employment of \gls{FR} systems is typically accompanied by prior deepfake detection systems\cite{Zhao2021Deepfake,qian2020deepfake} with the objective of detecting such tampered images. One popular approach in unlocking mobile phones is considering an additional infrared image, which makes it particularly challenging to create a suitable morphed face that matches the owner's infrared signature. However, if deepfake detection is not part of the \gls{FR} system or fails to detect the morphed faces, it is crucial to determine how susceptible state-of-the-art \gls{FR} systems are to such attacks.

Recent \gls{FR} systems \cite{arcface,deng2021variational,kim2020broadface} report face verification results exceeding 99.5\% on the arguably simple \gls{LFW} dataset \cite{LFW}, yet also reach 92\% under more challenging cross-age and cross-pose scenarios. 
First analyses of \gls{FR} performance under morphing attacks have been published before \cite{venkatesh2020influenceAgeAttacks,qin2021vulnerabilitiesPartialAttack,korshunov2019vulnerability,majumdar2019evading,zhang2021mipgan,venkatesh2020canMorph, scherhag2019facemorphinSurvey}. However, their results are limited as they only evaluate on frontal images with same gender and ethnicty \cite{venkatesh2020influenceAgeAttacks,zhang2021mipgan, qin2021vulnerabilitiesPartialAttack,majumdar2019evading,venkatesh2020canMorph,scherhag2019facemorphinSurvey}, only replace face parts \cite{qin2021vulnerabilitiesPartialAttack,majumdar2019evading}, 
 or do not evaluate state-of-the-art \gls{FR} methods \cite{venkatesh2020influenceAgeAttacks,qin2021vulnerabilitiesPartialAttack,korshunov2019vulnerability,majumdar2019evading,scherhag2019facemorphinSurvey}. To the best of our knowledge, no analysis has been published revealing the vulnerability of state-of-the-art \gls{FR} systems on challenging datasets comprising images taken in the wild. 
 
Our contributions can be summed up as follows:
\begin{itemize}
    \item We show how a pretrained \gls{FR} model can be employed for face morphing as an encoder in a \gls{GAN} with losses specifically adapted to face morphing. Our network gradually morphs two faces depending on a single parameter $\alpha$ yielding remarkable results.
    \item In our exhaustive analysis, which emphasizes on faces taken in the wild, we demonstrate how the accuracy of an \gls{FR} system is affected by morphed faces and how the knowledge of the \gls{FR} system influences the results. 
\end{itemize}


\section{Related Work}
\label{sec:relatedwork}
\subsection{Face Manipulation}

An early face manipulation approach used multi-scale inputs and supervised the generation with an additional illumination loss \cite{korshunova2017fast}. With the success of \glspl{GAN} in realistic image synthesis, a generator with an encoder-decoder structure is leveraged by the majority of the methods \cite{he2019AttGAN, Chen2019BeautyGlow, diamant2019BeholderGAN, Liu2021A3GAN, perov2021deepfacelab, nirkin2019fsgan,ngo2021selfManipul,wei2020maggan} to obtain photo-realistic results. To generate a face with the desired attributes, a conditional \gls{GAN} \cite{mirza2014CGAN} structure can be employed, in which the generator is provided with additional information and the discriminator also acts as a classifier. \Eg, Diamant \etal \cite{diamant2019BeholderGAN} incorporated a beauty score, while He \etal \cite{he2019AttGAN} used a binary attribute vector to guide the image synthesis in the decoder.

In face swapping or mapping the attributes of one face to the other face, the generator's input comprises two images. For this task, Chen \etal \cite{Chen2019BeautyGlow} disentangled makeup and non-makeup latent vectors to generate a new face containing the makeup of one face with the remaining properties (identity, background, attributes) of the other face. Nirkin \etal  \cite{nirkin2019fsgan} proposed a face swapping \gls{GAN}, which contains multiple generators for reenactment, segmentation, inpainting, and blending. In FaceShifter \cite{li2019faceshifter}, face identity and attribute features are extracted separately and induced into the decoder at different resolutions. The Mask-Guided \gls{GAN} \cite{wei2020maggan} further trains a mask to control the region where the features are modified. To combine face identity and face attribute features, the latter two approaches  \cite{li2019faceshifter,wei2020maggan} employ \gls{SPADE} \cite{park2019spade}, which renormalizes feature maps based on a learn transformation from the features. Ng\^{o} \etal \cite{ngo2021selfManipul} decomposed the face to adjust the head pose, light, and facial expression separately while maintaining identity and background information.

\begin{figure*}[t]
	\centering
	\includegraphics[width=\linewidth]{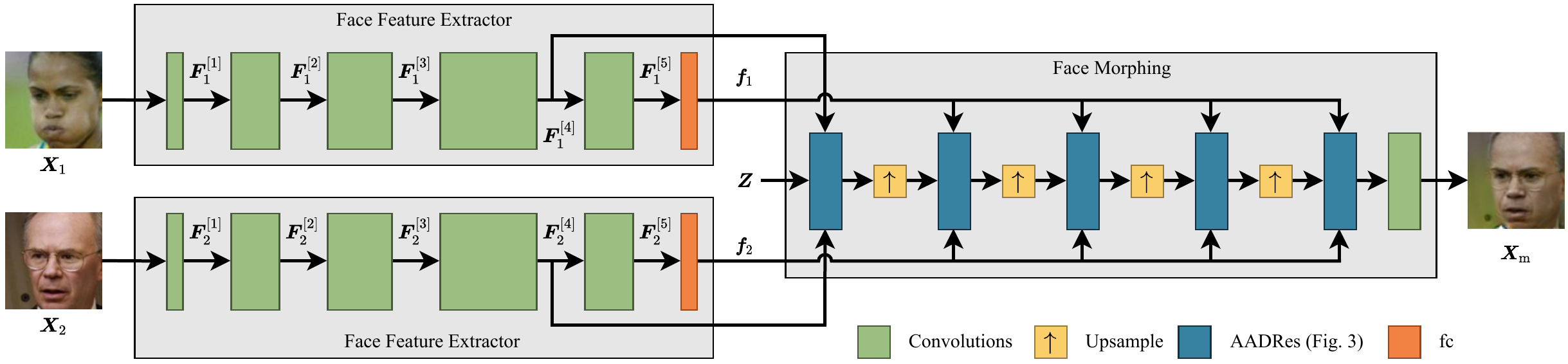}
	\caption{Our approach for face morphing: Face features of two faces  $\boldsymbol X_1$ and $\boldsymbol X_2$ are extracted by a ResNet-50. Then, the face morphing network utilizes face features $\boldsymbol{f}$ and a single feature map to transform a trainable weight $\boldsymbol Z$ into a morphed face  $\boldsymbol X_{\text{m}}$.}
	\label{fig:network}
\end{figure*}

\subsection{GAN Inversion}

The objective of \gls{GAN} inversion is to find the most accurate latent vector, which allows a pretrained \gls{GAN} to recover the input image. Then, by altering the latent vector, the face can be manipulated. Thus, in contrast to the face manipulation methods mentioned above, \gls{GAN} inversion only trains the encoder in the generator, whereas the decoder is a pretrained \gls{GAN}.

Recently, multiple approaches for image synthesis based on latent vectors have been proposed, among which the BigGAN \cite{brock2018biggan} and StyleGAN  \cite{karras2019style, karras2020stylegan2} are most popular. In both StyleGAN versions, Karras \etal \cite{karras2019style, karras2020stylegan2} incorporated \gls{AdaIn}\cite{huang2017adain,dumoulin2017adain} -- a similar mechanism to \gls{SPADE} \cite{park2019spade} -- to introduce the information of the so-called style vectors into the generator at multiple depths. With the employment of attention as first introduced by Self-Attention \gls{GAN} \cite{zhang2019sagan}, the realism and variety of the generated images were further improved \cite{brock2018biggan, daras2020yourlocalGAN}.

Optimization-based \gls{GAN} inversion methods \cite{abdal2019image2stylegan,abdal2020image2styleganPP} first select a random initial latent vector, which then is optimized through gradient descend to produce the desired output image. In their analyses, Abdal \etal \cite{abdal2019image2stylegan,abdal2020image2styleganPP} demonstrated many possibilities with impressive results, including even a smooth transition between two face halves of different identities \cite{abdal2020image2styleganPP}. Multi-Code \gls{GAN} \cite{gu2020multiCodeGAN} utilizes $N$ latent codes to generate $N$ intermediate feature maps, which are then combined, weighted by their adaptive channel importance scores, to recover the output image. In order to invert \glspl{GAN} comprising attention mechanism, Daras \etal \cite{daras2020yourlocalGAN} proposed to employ the discriminator's attention layer. After \gls{GAN} inversion, localized and semantic-aware edits can be performed by disentangling and clustering the semantic objects in activation maps \cite{collins2020editing} or leveraging SVMs in the latent space \cite{shen2020interpretingSVM,hormann2021face}. Venkatesh \etal \cite{venkatesh2020canMorph} applied the \gls{GAN} inversion technique from Image2StyleGAN \cite{abdal2019image2stylegan} to face morphing by averaging the latent vectors.

Learning-based \gls{GAN} inversion approaches train a separated encoder, which can be applied to all images and thus  dispenses with the need of applying backpropagation to obtain the corresponding latent vector for every image. Zhu \etal \cite{zhu2020Indomain} used a domain-guided encoder as a regularizer to preserve the latent vector within the semantic domain of the generator. The StyleGAN Encoder \cite{richardson2021encoding} builds an encoder to extract the feature maps of images and subsequently trains a mapping network to transform the feature maps into layer-specific style vectors, which control the image generation in StyleGAN. Based on the same principle, Xu \etal \cite{xu2021generative} introduced a spatial alignment module into the encoder structure to better capture the spatial information from the input image. By incorporating an iterative refinement mechanism, Alaluf \etal \cite{alaluf2021restyle} drew from the iterative manner of optimization-based methods while maintaining efficiency as no backpropagation is performed. Zhang \etal \cite{zhang2021mipgan} trained a ResNet-50 as a backbone to predict the latent vector and then obtain morphed faces by averaging the latent vectors corresponding to both input faces. Despite remarkable high-quality results, their analysis is restricted to frontal faces and not applicable to faces that are taken in the wild.



\section{Methodology}

\begin{figure}[!b]
	\centering
	\includegraphics[width=\linewidth]{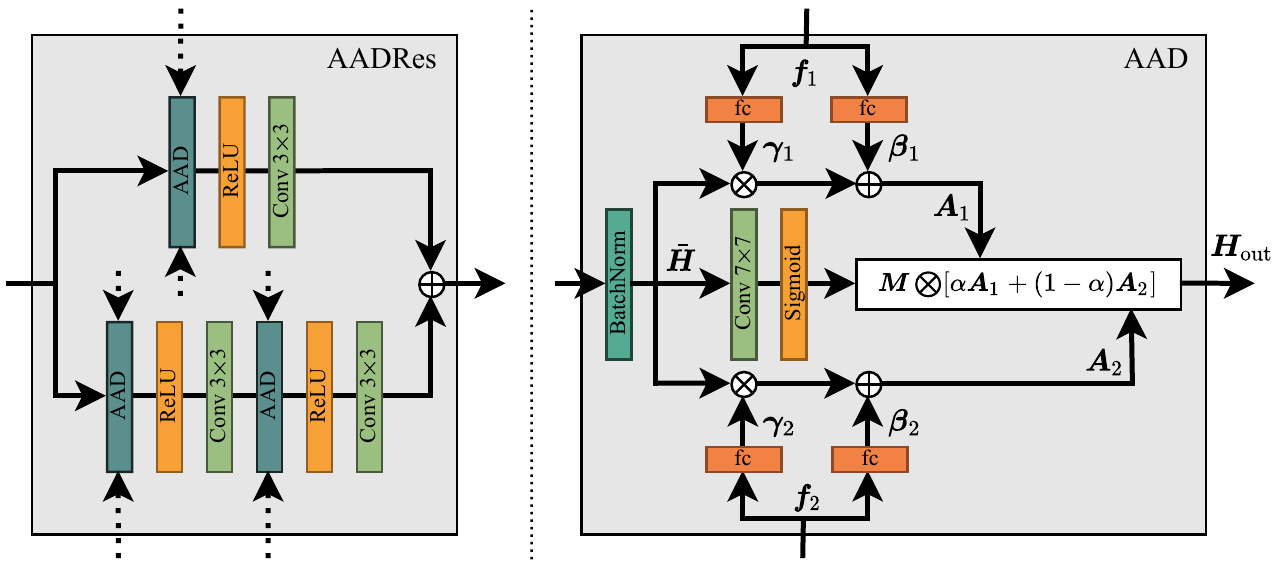}
	\caption{The elements of an \gls{AAD} residual block (left) following Li \etal \cite{li2019faceshifter} together with our modified \gls{AAD} block (right).}
	\label{fig:AAD}
\end{figure}

\subsection{Network Architecture}
As illustrated in \cref{sec:relatedwork}, there are two different architecture choices in training a face morphing network. We opted for a traditional approach not involving \gls{GAN} inversion. In this way, we can employ a pretrained \gls{FR} network to encode facial features from which the morphed face $\boldsymbol X_{\text{m}}$ is generated. This allows us to investigate how knowledge of the \gls{FR} system (white-box attack) affects the success rate of a face morphing attack. Due to the task definition, the network must further be invariant to the order of the inputs, \ie, assuming equal weights of $\boldsymbol X_1$ and $\boldsymbol X_2$, swapping the inputs must yield the same morphed face $\boldsymbol X_{\text{m}}$.

\cref{fig:network} depicts our approach to face morphing. The features $\boldsymbol f$ and intermediate feature maps $\boldsymbol F$ of two real faces $\boldsymbol X_1$ and $\boldsymbol X_2$ are extracted by a face feature extractor. Then, the morphed face $\boldsymbol X_{\text{m}}$ is generated by passing the previously extracted features (maps) through modified \glsfirst{AAD} residual blocks \cite{li2019faceshifter}.

\subsubsection{Feature Extractor}

The face feature extractor $E_{\text{gen}}(\cdot)$ is used to extract a representation $\boldsymbol f \in \mathbb{R}^{256}$ of the face. We use a ResNet-50 \cite{he2016identity}, which is pretrained on face identification tasks with softmax cross-entropy loss. Even though  additive angular margin (ArcFace \cite{arcface}) is widely employed nowadays, we decided to use a simpler approach for the face generation in order to demonstrate that even more powerful models, such as those trained with ArcFace, can be deceived by simpler approaches. 

\subsubsection{Face Morphing}
Our face morphing network is inspired by FaceShifter \cite{li2019faceshifter}, with several adaptions for face morphing.  The face morphing network starts with a trainable latent variable $\boldsymbol{Z}$ of size 7$\times$7$\times$512, which represents the initial feature maps and is deterministic in contrast to StyleGAN \cite{karras2019style}.

Then, the latent variable $\boldsymbol{Z}$ is propagated through a series of five \gls{AAD} residual blocks (\cf \cref{fig:AAD} (left)) with upsampling in between, similar to FaceShifter \cite{li2019faceshifter}. While the first \gls{AAD} residual block is fed with $\boldsymbol Z$ and the feature maps $\boldsymbol F_1^{[4]},\boldsymbol F_2^{[4]} \in \mathbb{R}^{7\times7\times1024}$, all subsequent blocks have the upsampled output of the previous blocks and the feature vectors $\boldsymbol f_1,\boldsymbol f_1$ as inputs. Moreover, the first \gls{AAD} residual block maintains the number of feature maps and therefore uses a skip connection as the upper path, whereas the remaining \gls{AAD} residual blocks halve the number of feature maps to reach 32 after the last one. We resize the feature maps with bilinear interpolation as upsampling to reduce checkerboard artifacts, which are frequently introduced by transposed convolutions \cite{odena2016dseconvolution}. Then, the face morphing network is concluded with a 3$\times$3 and 1$\times$1 convolution followed by clipping to obtain the output with the same dimensions and value range as the inputs. 

The crucial component of the face morphing network is the \gls{AAD} block, as illustrated by \cref{fig:AAD} (right) for $\boldsymbol f$ as input. First, the input feature map $\boldsymbol H$ is normalized with a batch normalization layer yielding $\bar{\boldsymbol H}$. Based on the normalized input $\bar{\boldsymbol H}$, a convolutional layer with sigmoid activation is employed to compute a mask $\boldsymbol{M}$, which indicates the activations in the feature maps to be changed within the \gls{AAD} block. Besides the mask prediction, every feature map of the normalized input $\bar{\boldsymbol H}$ is de-normalized yielding
\begin{equation}
    \boldsymbol A = \boldsymbol \gamma \bar{\boldsymbol H} + \boldsymbol \beta,
\end{equation}
with the target mean $\boldsymbol \beta$ and variance $\boldsymbol \gamma$, which are obtained by passing $\boldsymbol f$ through a fully connected layer, whose number of neurons match the number of feature maps of $\bar{\boldsymbol H}$.

In the first \gls{AAD} residual block with $\boldsymbol F^{[4]}$ as input, the input is flattened before applying the fully connected layer. We found that using $\boldsymbol F^{[4]}$ instead of $\boldsymbol f$ is crucial to obtain the smooth transition of both images as additional rough spatial information is provided through  $\boldsymbol F^{[4]}$. 

Information about the faces $\boldsymbol X_1$ and $\boldsymbol X_2$ are encoded in $\boldsymbol A_1$ and $\boldsymbol A_1$ as distinct de-normalizations of $\bar{\boldsymbol H}$. Unlike in FaceShifter \cite{li2019faceshifter}, we want to smoothly transition between the faces $\boldsymbol X_1$ and $\boldsymbol X_2$. Hence, we define a scalar parameter $\alpha \in [0;1]$, which globally balances the influence from $\boldsymbol A_1$ and $\boldsymbol A_2$ in every \gls{AAD} block. Then, the output $\boldsymbol H_{\text{out}}$ constitutes the element-wise multiplication of the mask $\boldsymbol M$ with the balanced encoded face features 
\begin{equation}
    \boldsymbol H_{\text{out}} = \boldsymbol M \otimes[\alpha\boldsymbol A_1 + (1-\alpha) \boldsymbol A_2].
\end{equation}

With our modifications to the original \gls{AAD} block \cite{li2019faceshifter}, we obtain invariance with respect to $\boldsymbol X_1$ and $\boldsymbol X_2$ by design. This is achieved by sharing the weights of the face feature extractor and the fully connected layers within every \gls{AAD} block, which are used to compute $\boldsymbol \gamma$ and $\boldsymbol \beta$. Moreover, every \gls{AAD} decides with its own mask $\boldsymbol M$ which values of the current feature map to manipulate. 

\subsection{Loss Functions}

To train our face morphing model, we use a weighted sum of several losses
\begin{equation}  
    \mathscr{L}_{\text{G}}= \lambda_{\text{adv}}\mathscr{L}_{\text{adv}}^{\text{G}} + \lambda_{\text{id}}\mathscr{L}_{\text{id}}+ \lambda_{\text{per}}\mathscr{L}_{\text{per}}+\lambda_{\text{style}}\mathscr{L}_{\text{style}},
\end{equation}
where $\lambda_{\text{adv}}$, $\lambda_{\text{id}}$, $\lambda_{\text{per}}$, and $\lambda_{\text{style}}$  denote scalars used to balance the losses.

Similar to most image manipulation approaches in which realism plays a vital role, we utilize the face morphing network as a generator in a \gls{GAN} structure and train with an adversarial loss. In this way, the face morphing network must generate a photo-realistic face to deceive the discriminator, whereas the discriminator tries to discern real faces $\boldsymbol{X}_{1}$ or $\boldsymbol{X}_{2}$ from the synthetically generated morphed face $\boldsymbol{X}_{\text{m}}$. We implement a global discriminator $D(\boldsymbol X)$ comprising four convolutional layers -- the first two with stride two, which are concluded by a fully connected layer and sigmoid activation function denoting the probability of the input image $\boldsymbol X$ being real. Then, the adversarial losses are  

\begin{equation}
    \mathscr{L}_{\text{adv}}^{\text{G}} = -\log(D(\boldsymbol{X}_{\text{m}})),
\end{equation}
\begin{equation}
    \mathscr{L}_{\text{adv}}^{\text{D}} = -\log(1-D(\boldsymbol{X}_{\text{m}})) - \frac{1}{2} \sum\limits_{i=1}^2\log(D(\boldsymbol{X}_{i})).
\end{equation}

Similar to \cite{li2019faceshifter,venkatesh2020canMorph,zhang2021mipgan}, we employ an identity loss, which forces the face morphing network to generate the face $\boldsymbol X_{\text{m}}$ that matches $\boldsymbol{X}_{1}$ and $\boldsymbol{X}_{2}$. The parameter $\alpha$ indicates how much information the \gls{AAD} block utilizes from $\boldsymbol{X}_{1}$ compared to $\boldsymbol{X}_{2}$. Thus, this influence is also reflected in the identity loss
\begin{align}
    \mathscr{L}_{\text{id}} &= \alpha d_{\text{cos}}\left(\boldsymbol{f}_{\text{m}}, \boldsymbol{f}_{1} \right) + (1-\alpha) d_{\text{cos}}\left(\boldsymbol{f}_{\text{m}}, \boldsymbol{f}_{2} \right),
\end{align}
where $d_{\text{cos}}(\cdot,\cdot)$ denotes the cosine distance between two feature vectors. To further guide the face morphing network to output a face $\boldsymbol{X}_{\text{m}}$ containing information from both inputs, we adapt the perceptual $\mathscr{L}_{\text{per}}$ and style loss $\mathscr{L}_{\text{style}}$ from Johnson \etal \cite{johnson2016perceptual} by incorporating $\alpha$
\begin{align}
\mathscr{L}_{\text{per}}=\sum_{i=4}^{5} \frac{\alpha}{N^{[i]}}\left\lVert\boldsymbol{F}_{1}^{[i]}-\boldsymbol{F}_{\text{m }}^{[i]}\right\rVert_1 +  \frac{(1-{\alpha})}{N^{[i]}}\left\lVert\boldsymbol{F}_{2}^{[i]}-\boldsymbol{F}_{\text{m}}^{[i]}\right\rVert_1,
\end{align}
with $N^{[i]}$ being the number of elements in $\boldsymbol{F}^{[i]}$.  Besides the adversarial loss to ensure photo-realistic results, perceptual loss \cite{johnson2016perceptual} is widely employed to ensure matching feature maps \cite{Chen2019BeautyGlow,diamant2019BeholderGAN,Liu2021A3GAN, nirkin2019fsgan, ngo2021selfManipul, xu2021generative, richardson2021encoding, gu2020multiCodeGAN}. 

The style loss $\mathscr{L}_{\text{style}}$ uses the Gram matrix of every feature map and was modified from \cite{johnson2016perceptual} accordingly. Both latter losses ensure that the transition of $\boldsymbol{X}_{\text{m}}$ from $\boldsymbol{X}_{1}$ to $\boldsymbol{X}_{2}$ is visible in the feature maps. In contrast to other works on face morphing \cite{venkatesh2020canMorph,zhang2021mipgan}, we   decided to compute $L_{\text{per}}$ and $\mathscr{L}_{\text{style}}$ based on rather deep feature maps $\boldsymbol{F}^{[4]}$ and $\boldsymbol{F}^{[5]}$ as they do contain less spatial information and thus less ambiguity, \ie, the network is not forced to generate two noses if their corresponding activations are at different locations in shallower feature maps. Moreover, we employ a feature extractor trained on faces to increase the meaningfulness of such feature maps.

\section{Experiments}
\label{sec:experiments}

\subsection{Training Details}
\label{sec:training}

To demonstrate that the morphed face not only deceives the feature extractor used for face morphing $E_{\text{gen}}(\cdot)$, we also utilize a more sophisticated feature extractor trained with ArcFace loss $E_{\text{arc}}(\cdot)$ and apply it on a different dataset. Both feature extractors were trained with facial images of size 112$\times$112, which were aligned utilizing the landmarks obtained by MTCNN \cite{zhang2016joint}. While we use VGGFace2 to train $E_{\text{gen}}(\cdot)$, the refined MS-Celeb-1M was utilized for $E_{\text{arc}}(\cdot)$ \cite{MS1M,arcface}.

Next, the weights of $E_{\text{gen}}(\cdot)$ are fixed and the face morphing network is trained with $\mathscr{L}_{\text{G}}$ ($\lambda_{\text{adv}}=1$, $\lambda_{\text{id}}=2$, $\lambda_{\text{per}}=0.5$, and $\lambda_{\text{style}}=120$) in an alternating manner with the discriminator $\mathscr{L}_{\text{adv}}^{\text{D}}$. To ease convergence, we first train with $\boldsymbol X_1 = \boldsymbol X_2$ for 5 epochs and finetune with $\boldsymbol X_1 \neq \boldsymbol X_2$ for another 10 epochs using Adam optimizer. Every batch comprises 32 faces from exactly 16 different identities. We found that the face morphing network improves very slowly and thus only use every fourth batch to train the discriminator. Moreover, the learning rates of the face morphing network and the discriminator are set to $10^{-4}$ and $10^{-5}$, respectively. Both are decayed by a factor of 0.5 every third epoch. For the parameter $\alpha$, we implement two versions: 1) fixed at $\alpha=0.5$ throughout the training; and 2) $\alpha=0.5$ during pretraining and a truncated Gaussian distribution with mean $\mu=0.5$ and variance $\sigma=0.2$ for finetuning to ensure that also faces morphed with $\alpha \neq 0.5$ are realistic.

\subsection{Benchmark Details}
\label{sec:benchDetails}
Typical benchmarks for face verification can be seen as a list of triplets $\mathcal{T}=\left(\boldsymbol X_1, \boldsymbol X_2, y \right)$ with $y=1$ denoting that $\boldsymbol X_1$ and $\boldsymbol X_2$ have the same identity ($\id(\boldsymbol X_1)=\id(\boldsymbol X_2)$) and $y=0$ if not. For our task, we want to obtain the accuracy $Acc_{\text{morph}}$ of a face feature extractor $E_{\text{test}}(\cdot)$ correctly classifying $\boldsymbol{X}_{\text{m}}$ if $\boldsymbol{X}_{\text{m}}$ was generated from two faces with different identities, \ie, the desired classification is $\id(\boldsymbol X_{\text{m}})\neq\id(\boldsymbol X_1)$ and $\id(\boldsymbol X_{\text{m}})\neq\id(\boldsymbol X_2)$. Formally, $Acc_{\text{morph}}$ is computed as
\begin{align}
    Acc_{\text{morph}} = 1- \frac{1}{N_{\text{diff}}}|\{\forall~\mathcal{T} \mid~&  d_{\text{cos}}\left(\boldsymbol{f}_{\text{m}}, \boldsymbol{f}_{1} \right) < t ~\&~y = 0~\&\nonumber\\
    &d_{\text{cos}}\left(\boldsymbol{f}_{\text{m}}, \boldsymbol{f}_{2} \right) < t\}|,
\label{eq:acc}
\end{align}
with $N_{\text{diff}}$ denoting the number of imposter pairs ($y=0$) and $t$ the threshold, which is computed to maximize the traditional accuracy of the respective verification protocol. Thus, $Acc_{\text{morph}}$ can also be referred to as the failure rate of a face morphing attack onto a \gls{FR} system. We further compute \cref{eq:acc} for $y=1$, \ie, genuine pairs, to affirm that the morphed face generated by using two faces from the same identity is perceived as another image of that identity. 

Since we have designed our face morphing network to allow a gradual change from $\boldsymbol X_1$ and $\boldsymbol X_2$, $\boldsymbol X_{\text{m}}$ always looks similar to both input faces $\boldsymbol X_1$ and $\boldsymbol X_2$. Therefore, we extended the triplets in the benchmark to quintuples by adding two images $\boldsymbol X_1'$ and $\boldsymbol X_2'$, which match the identities of $\boldsymbol X_1$ and $\boldsymbol X_2$, respectively.\footnote{The protocol is available under: \url{https://github.com/stefhoer/FaceMorph}} Thus, $\id(\boldsymbol X_1)=\id(\boldsymbol X_1')$ and $\id(\boldsymbol X_2)=\id(\boldsymbol X_2')$. Then, $\boldsymbol X_{\text{m}}$ is still created based on $\boldsymbol X_1$ and $\boldsymbol X_2$, yet the features of $\boldsymbol X_1'$ and $\boldsymbol X_2'$ are used for evaluation and to compute the threshold $t$. This quintuple protocol is denoted by $\dagger$ in our analysis.

For our analysis, we use the \gls{LFW} \cite{LFW} dataset together with the cross-age and cross-pose extensions CALFW \cite{CALFW} and CPLFW \cite{CPLFW}. All three benchmark datasets contain 6000 pairs (3000 imposter and 3000 genuine pairs) and are evaluated using 10-fold cross-validation. Even though nowadays the \gls{LFW} dataset is not very helpful in evaluating face verification accuracy due to its obvious imposter pairs, it fits our purpose since other datasets ensure that imposter pairs have the same gender and ethnicity, which eases deceiving the \gls{FR} system. By using $\boldsymbol X_1$ and $\boldsymbol X_2$ to generate $\boldsymbol X_{\text{m}}$, we maintain the properties (cross-age and cross-pose) of the CALFW and CPLFW dataset, whereas the threshold $t$ computation is based on $\boldsymbol X_1'$ and $\boldsymbol X_2'$ to ensure that the same threshold is used to distinguish $\boldsymbol X_1'$ from $\boldsymbol X_2'$, and $\boldsymbol X_{\text{m}}$ from $\boldsymbol X_1'$, $\boldsymbol X_2'$. However, our method does not guarantee an age or a pose difference between the newly selected $\boldsymbol X_1'$ and $\boldsymbol X_2'$ as in the original CALFW and CPLFW benchmarks. Still, the same gender and ethnicity are maintained as defined in the original protocols. 

\begin{figure}[t]
	\centering
    \setlength\figlen{3.8cm}
	\input{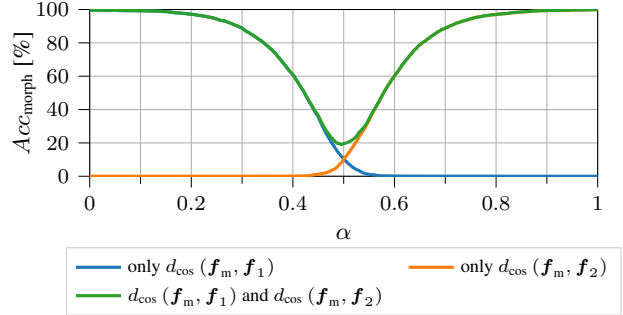}
 	\vspace{-0.4cm}
	\caption{The accuracy $Acc_{\text{morph}}$ of ArcFace for different $\alpha$ when distinguishing faces morphed by the model trained with Gaussian-distributed $\alpha$ and without $\mathscr{L}_{\text{style}}$. Faces are morphed based on the \gls{LFW} benchmark with imposter pairs and $Acc_{\text{morph}}$ is computed also separately for every $\boldsymbol X_1$ and $\boldsymbol X_2$.}
	\label{fig:varyingAlpha}
 	\vspace{-0.3cm}
\end{figure} 

\begin{table*}[t]
  \centering
  \caption{Ablation study: Accuracy $Acc_{\text{morph}}$ [\%] of the \gls{FR} system $E_{\text{test}}(\cdot)$ when classifying morphed faces $\boldsymbol{X}_{\text{m}}$ on \gls{LFW}, CALFW, and CPLFW datasets. \textit{Same} and \textit{diff} denote whether the input images have the same identity and $\dagger$ indicates that different images were used for evaluation than for morphing.}
    \resizebox{\linewidth}{!}{\begin{tabular}{lcc|cccccccccc}
    \toprule
    &&&\multicolumn{6}{c}{LFW}& \multicolumn{2}{c}{CALFW} & \multicolumn{2}{c}{CPLFW}\\
    \cmidrule(lr){4-9}  \cmidrule(lr){10-11} \cmidrule(lr){12-13}
          &       &       & \multicolumn{2}{c}{$E_{\text{test}}(\cdot) = E_{\text{gen}}(\cdot)$} & \multicolumn{4}{c}{$E_{\text{test}}(\cdot) = E_{\text{arc}}(\cdot)$}& \multicolumn{4}{c}{$E_{\text{test}}(\cdot) = E_{\text{arc}}(\cdot)$} \\
          \cmidrule(lr){4-5}          \cmidrule(lr){6-9} \cmidrule(lr){10-13}
    \multirow[t]{2}{*}{\shortstack[l]{Features for\\ face morphing}}  & $\mathscr{L}_{\text{style}}$ & $\alpha$ & same  & diff  & same  & diff  & same $\dagger$ & diff $\dagger$ & diff & diff $\dagger$ & diff & diff $\dagger$ \\
    \midrule
    $\boldsymbol f$ & $\surd$   & 0.5 & \textbf{0.0} & 1.1  & 0.2  & 32.2  & 2.1  & 72.9  & 19.0 & 71.7 & 19.1 & 73.9 \\
    $\boldsymbol f,\boldsymbol F^{[3]},\boldsymbol F^{[4]}$ & $\surd$   & 0.5 & \textbf{0.0} & 0.5  & 0.1  & 19.7  & 0.8  & 62.8  & 7.7 & 58.9 & 19.2 & 71.4 \\
    $\boldsymbol f,\boldsymbol F^{[4]}$ & $\surd$   & 0.5 & \textbf{0.0} & \textbf{0.0} & 0.1  & \textbf{18.2} & 0.9  & 62.7  & 6.4  & 59.4 & \textbf{15.3} & 68.3 \\
    $\boldsymbol f,\boldsymbol F^{[4]}$ &   &   0.5     & \textbf{0.0} & \textbf{0.0} & 0.1  & 19.2  & 0.9  & 62.2  & 6.4 & 58.5 & 16.5 & \textbf{68.2} \\
    $\boldsymbol f,\boldsymbol F^{[4]}$ & $\surd$  & $\mathcal{N}(0.5,0.2)$  & \textbf{0.0} & 4.3  & \textbf{0.0} & 19.6  & \textbf{0.6} & \textbf{60.8} & 5.2 & 54.5 & 16.2 & 69.7 \\
    $\boldsymbol f,\boldsymbol F^{[4]}$ &          & $\mathcal{N}(0.5,0.2)$ &  0.1 & 0.8 & 0.1  & 19.5  & 0.7  & 61.1  & \textbf{4.7}  & \textbf{54.2} & 16.1 & 68.7 \\
    \bottomrule
\end{tabular}}%
  \label{tab:ablation}%
 	\vspace{-0.3cm}
\end{table*}%

The extension from triplets to quintuples requires modifications to the original \gls{LFW} protocol as many identities in the \gls{LFW} dataset only have a single image. 1165 genuine and 2736 imposter pairs were replaced, reducing the number of identities covered by the benchmark from 3158 to 1648. For CALFW and CPLFW, no pairs were substituted as at least two images per identity were available. Despite the inherent reduction of generalization due to fewer identities in the quintuples \gls{LFW}, the frequent differences of ethnicity and gender in imposter pairs still render it particularly interesting.

\subsection{Quantitative Results}

\cref{fig:varyingAlpha} illustrates the change in accuracy $Acc_{\text{morph}}$ as introduced in \cref{eq:acc} together with the accuracies per identity. It is apparent that our approach provides a smooth transition between two faces in the feature space $\boldsymbol f$. While $\alpha \approx 0$ results in a morphed face $\boldsymbol X_{\text{m}}$, which is never classified to have the same identity as $\boldsymbol X_1$ resulting in an accuracy close to 100\%, it contains enough information to be classified as $\id(\boldsymbol X_2)$ and fool the system. For $\alpha \approx 1$, this behavior is inverted. The lowest accuracy of rejecting $\boldsymbol X_{\text{m}}$ for at least one identity $Acc_{\text{morph}}=19.5$\% is achieved for $\alpha\approx0.5$, where the information from $\boldsymbol X_1$ and $\boldsymbol X_2$ to generate  $\boldsymbol X_{\text{m}}$ is considered equally. Still, the network classifies $\boldsymbol X_{\text{m}}$ as either $\id(\boldsymbol X_1)$ or $\id(\boldsymbol X_2)$ in 89.8\% of the cases.

\cref{tab:ablation} depicts the results of our approach for different configurations. When considering genuine pairs (same), it is evident that the morphed face $\boldsymbol X_{\text{m}}$ is always classified as the identity  -- even in the more challenging scenario when the faces used for morphing and evaluation differ ($\dagger$). 

In the more applicable scenario of morphing faces from imposter pairs (diff), the differences between the configurations become apparent. Morphing a face solely based on two $256$-dimensional identity vectors $\boldsymbol f$ yields inferior results on all protocols. This demonstrates that the spatial information present in $\boldsymbol F^{[4]}$ is crucial for achieving satisfying results. Incorporating feature maps $\boldsymbol F^{[3]}$ with a resolution of 14$\times$14 leads to worse results. We conjecture that including $\boldsymbol F^{[3]}$ confuses the network in many cases as information in $\boldsymbol F^{[3]}$ is more ambiguous due to the larger resolution.

According to the results reported in \cref{tab:ablation}, utilizing style loss $\mathscr{L}_{\text{style}}$ in addition to the perceptual loss $\mathscr{L}_{\text{per}}$ cannot be considered beneficial. Independent of $\alpha$, not employing $\mathscr{L}_{\text{style}}$ results in a slight improvement, which becomes more noticeable on the more challenging and relevant cases with quintuples ($\dagger$). Gaussian-distributed $\alpha$ lowers $Acc_{\text{morph}}$ most noticeably on CALFW.

\begin{figure}[t]
	\centering
    \setlength\figlen{3.55cm}
	\input{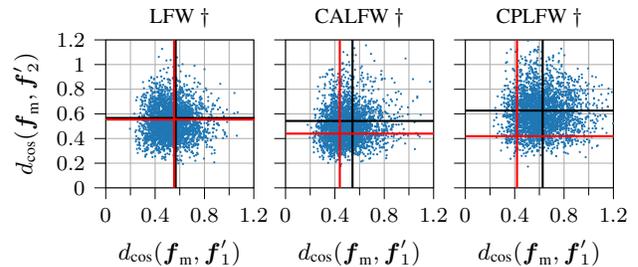}
 	\vspace{-0.5cm}
	\caption{Cosine feature distances between $\boldsymbol X_{\text{m}}$, generated by the model trained with Gaussian-distributed $\alpha$ and without $\mathscr{L}_{\text{style}}$, and $\boldsymbol X_1'$, $\boldsymbol X_2'$ extracted with ArcFace. Operating thresholds maximizing the traditional accuracy (black) or obtaining a false accept rate $FAR=0.1$\% (red) indicate the decision boundaries. Thus, distances lying in the bottom left quadrant were misclassified.}
	\label{fig:scatter}
 	\vspace{-0.3cm}
\end{figure} 

\begin{figure*}[t]
	\centering
	\includegraphics[width=\linewidth]{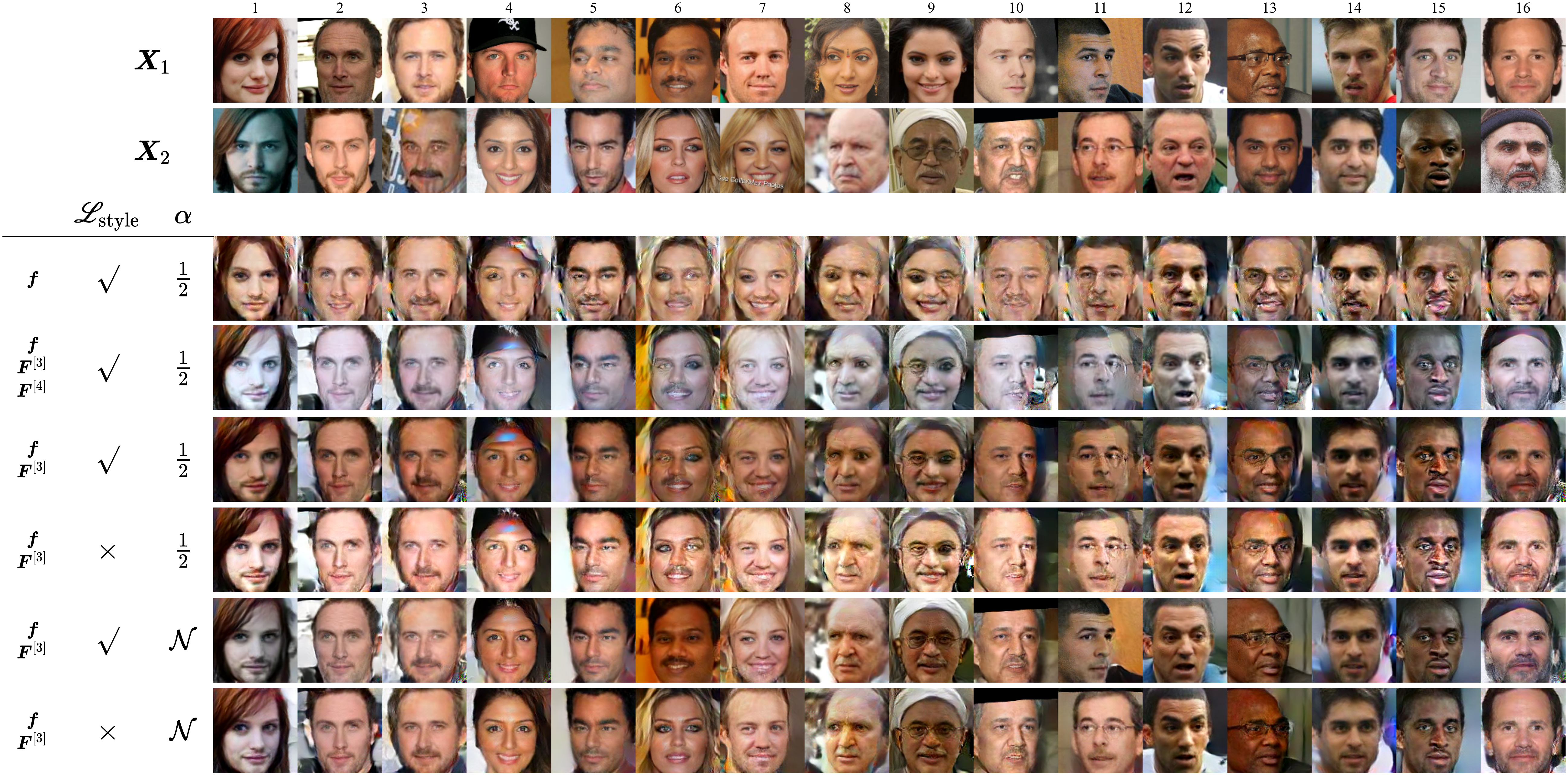}
	\caption{Morphed faces $\boldsymbol X_{\text{m}}$ generated from $\boldsymbol X_1$ and $\boldsymbol X_2$
	by models trained with different parameters following \cref{tab:ablation}. $\boldsymbol X_1$ and $\boldsymbol X_2$ are selected from LFW dataset and $\boldsymbol X_{\text{m}}$ is computed for $\alpha = 0.5$.}
	\label{fig:qual_results}
 	\vspace{-0.3cm}
\end{figure*}
\begin{figure*}[t]
	\centering
	\includegraphics[width=\linewidth]{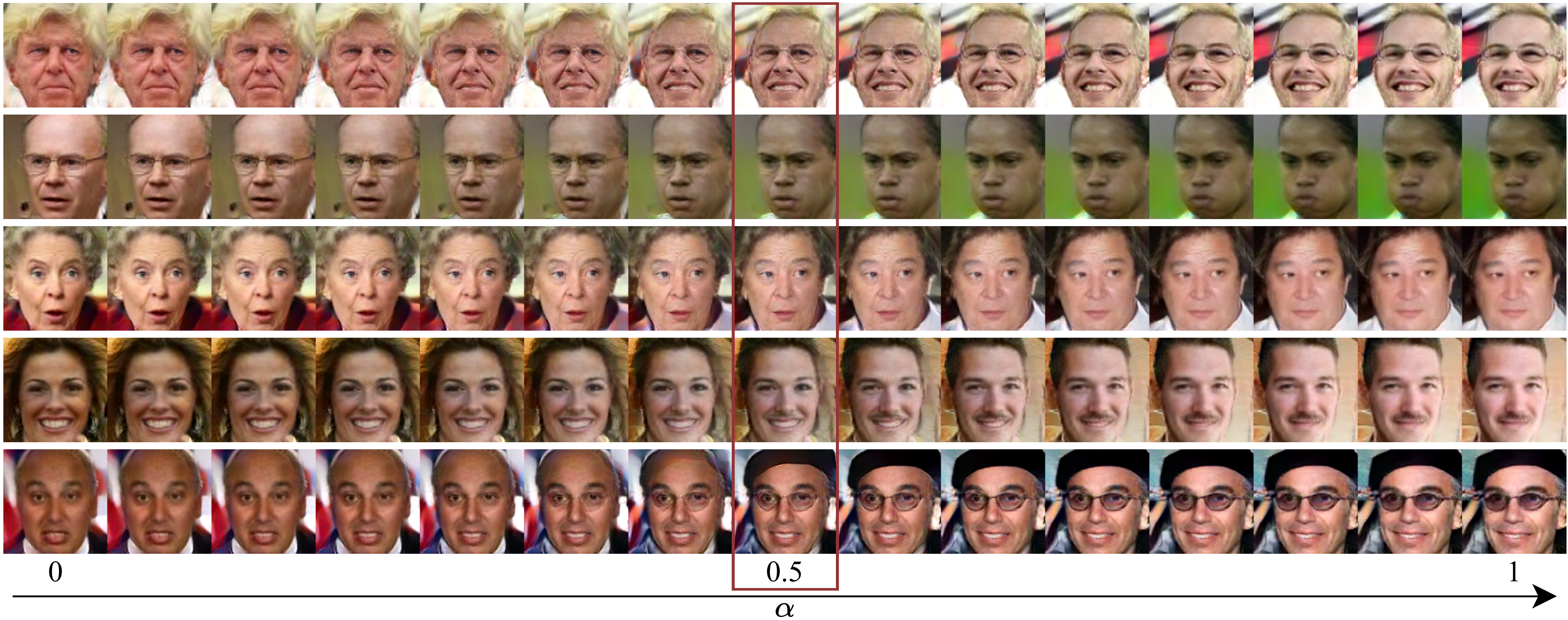}
	\caption{Gradual change of morphed faces $\boldsymbol X_{\text{m}}$ generated by the model trained with Gaussian-distributed $\alpha$ and without $\mathscr{L}_{\text{style}}$.}
	\label{fig:qual_results_gradual}
 	\vspace{-0.3cm}
 \end{figure*}

The inferior accuracies $Acc_{\text{morph}}$ on CALFW compared to \gls{LFW} confirm the suitability of \gls{LFW} for this analysis since imposter pairs in CALFW were selected to have the same gender and ethnicity, which facilitates face morphing. Even when morphing faces with large head poses variations as in CPLFW, the face morphing network deceives $E_{\text{arc}}(\cdot)$ with a success rate of over 30\%. Still, varying head poses represent one of the biggest challenges in face morphing, which is also affirmed by visual inspection of the feature distances in \cref{fig:scatter}. When using an operating threshold typical in security-sensitive applications corresponding to a false accept rate $FAR=0.1$\%, only 1.3\% of the morphed faces can fool the \gls{FR} system. Still, one must note that this behavior is expected in challenging scenarios and leads to relatively low true accept rates rendering the whole system impractical.

Our analysis also highlights the dependency on the evaluation protocol. When considering a white-box attack, \ie, using the same feature extractor for morphing and evaluation $E_{\text{test}}(\cdot) = E_{\text{gen}}(\cdot)$, the face morphing network learns to fool this specific system limiting $Acc_{\text{morph}}$ to $\approx0$\% for most models, even if different identities are morphed. However, if a more sophisticated model $E_{\text{arc}}(\cdot)$ is employed for testing, the \gls{FR} system detects morphed faces with $Acc_{\text{morph}} >18$\%. This is reasonable as the morphed face was not generated to deceive $E_{\text{arc}}(\cdot)$, which focuses on different face features due to its distinct loss and training dataset. For the most practical and challenging scenario of using different faces for morphing and evaluation ($\dagger$), the accuracy $Acc_{\text{morph}}$ exceeds 60\%. Nevertheless, being able to fool the system in 30-45\% of the cases, depending on the gender and pose of both faces, clearly demonstrates the susceptibility of state-of-the-art methods to face morphing. Besides, if one would want to fool a system using frontal faces of similar ethnicity and gender would be an obvious choice, which results in a success rate of up to $45.8$\%.

\subsection{Qualitative Results}

Many morphed faces $\boldsymbol X_{\text{m}}$ in \cref{fig:qual_results} display noticeable artifacts, making it easy for a human to spot that the face must have been manipulated. Still, with the rise of automatic border control or automatic access systems, missing realism is only a small disadvantage if plausibility checks are not employed. Moreover, the last two rows, \ie, employing Gaussian-distributed $\alpha$ during training, showed the most realistic results in accordance with \cref{tab:ablation}. Particularly interesting results are further shown by the model only provided with features $f_1$ and $f_2$. Here, the absence of spatial information causes the model to generate always frontalized $\boldsymbol X_{\text{m}}$, which further demonstrates that certain information such as accessories are not encoded into $f_1$ and $f_2$ in the first place. \cref{fig:qual_results_gradual} visually confirms the quantitative analysis in \cref{fig:varyingAlpha} in that our face morphing network achieves a seamless change between two faces.

\section{Conclusion}

This paper presents a method of using an existing pretrained \gls{FR} model to generate morphed faces. The \gls{FR} model is used to extract face identity features and feature maps, which guide the decoder in generating a morphed face. By adapting the \gls{AAD} block and multiple losses to face morphing, we achieve a seamless transition between two faces -- visually and in the feature space. Compared to previous approaches, we also encompass pairs of faces with varying head poses, different gender, or ethnicity. Our exhaustive analysis demonstrates that state-of-the-art \gls{FR} are vulnerable to morphed faces even if a relatively simple \gls{FR} model is employed to generate the morphed face. Besides, we analyze the influence of knowing the \gls{FR} model (white-box attack) and show that morphed faces with extreme head poses are less likely to be misclassified. Overall, our work highlights the necessity of using deepfake detection -- particularly when employing \gls{FR} in security-sensitive scenarios.

{\small
\bibliographystyle{ieee_fullname}
\bibliography{egbib}
}

\end{document}